\documentclass[10pt,twocolumn]{article}
\pdfoutput=1
\usepackage{cite}
\usepackage[T1]{fontenc}
\usepackage{hyperref}
\usepackage{url}
\usepackage{microtype}
\usepackage{mdframed}
\usepackage{dcolumn}
\usepackage{graphicx}
\usepackage{xcolor}
\usepackage[font=small,labelfont=bf]{caption}
\usepackage{color,soul}
\usepackage{geometry}
\usepackage{authblk}
\usepackage{float}

\hypersetup{pdfauthor={Marvell and Nokia},pdftitle={Pre-Quantized Deep Learning Models Codified in ONNX to Enable Hardware/Software Co-Design}}

\definecolor{darkgreen}{rgb}{0.0, 0.42, 0.24}

\usepackage{listings}

\begin{document}

\title{\vspace{-1em}\huge{\textbf{Pre-Quantized Deep Learning Models Codified in ONNX to Enable Hardware/Software Co-Design}}}
\author[1]{Ulf Hanebutte}
\author[2]{Andrew Baldwin}
\author[1]{Senad Durakovic}
\author[3]{Igor Filipovich}
\author[1]{Chien-Chun (Joe) Chou}
\author[4]{Damian Adamowicz}
\author[1]{Derek Chickles}
\author[5]{David Hawkes}

\affil[1]{Marvell Technology Inc.}
\affil[2]{Nokia Solutions and Networks Oy}
\affil[3]{Nokia of America Corporation}
\affil[4]{Nokia Solutions and Networks Sp. z o.o.}
\affil[5]{Nokia Paris Saclay}

\date{}
\maketitle

\begin{abstract}
This paper presents a methodology to separate the quantization process from the hardware-specific model compilation stage via a pre-quantized deep learning model description in standard ONNX format. Separating the quantization process from the model compilation stage enables independent development. The methodology is expressive to convey hardware-specific operations and to embed key quantization parameters into a ONNX model which enables hardware/software co-design. Detailed examples are given for both MLP and CNN based networks, which can be extended to other networks in a straightforward fashion.
\end{abstract}

\section{Introduction}
\label{sec:intro}

In recent years, deep learning models have shown to provide superior predictive capabilities in many domains. However, these models, as they are typically developed in full $fp32$ floating-point precision, are both extreme in compute and memory intensive. The mathematical concept of quantization which generates models in lower precision, e.g. $int8$, has gained importance as a way to alleviate the computational requirements. In particular, quantization is a critical step in generating hardware-platform optimized models of today’s deep learning accelerators.

Typically, models are developed as full-precision ($fp32$) models and the quantization is part of the compilation flow. Pre-quantized models are models that are already quantized by the time the models are passed to compilation.

Why is this important for co-design?
The ability to separate the quantization process from the hardware platform-specific compilation stage is important for optimal deep learning model execution. It allows researchers and modeling toolchain developers to focus on the quantization process independent of the specific hardware-platform. Thus, researchers/developer are able to create domain specific pre-quantized models, rather than rely on a general-purpose quantization approach that a hardware platform-specific compilation flow provides. However, to utilize hardware architecture specific features, the description of a pre-quantized model needs to be expressive. The design of this separated quantization process has the following goals:

\noindent
\begin{enumerate}
\itemsep0em
\item{Key quantization parameters should be embedded into the ONNX model.}
\begin{itemize}
\item {No additional target-specific external metadata must be required.}
\end{itemize}
\item{Model should be directly executable with standard ONNX tools, such as ONNXruntime ~\cite{onnxruntime}.}
\item{Standardized ONNX ~\cite{onnx} operators should only be used.}
\begin{itemize}
\item {No custom operators which would prevent model usage in standard tools.}
\item {Closely matching output (within narrow margins) on all inference environments (software or hardware).}
\end{itemize}
\item{The description of the model should be expressive to convey hardware-specific operations.}
\begin{itemize}
\item {E.g. codify integer scale and right bit shift used by hardware to perform rescaling.}
\end{itemize}
\end{enumerate}
In the following section, the paper provides an introduction to symmetric quantization, followed by detailed examples of ONNX representation of pre-quantized Fully Connected Layer and Convolution Layers. This methodology is further applied to Tanh and Sigmoid activation functions.

\section{Related Work}
\label{sec:related}
As quantization has gained importance, many deep learning frameworks and compilers have implemented some form of quantization. Below we cite related work. It should be noted, that in general the cited work is focused on quantized networks within their frameworks and compile chain. However, they are not addressing the need to codify an already quantized, i.e., a pre-quantized model, in a standard format. We cite here:

\noindent
\begin{itemize}
\item TensorFlow lite ~\cite{tf_lite}
\item Nvidia\textsuperscript{\textregistered} TensorRT\textsuperscript{TM}~\cite{tensorRT}
\item Onnxruntime how to quantize ~\cite{onnxruntime_howto_quantize}
\item Profile-guided quantization in Glow ~\cite{rotem2019glow}
\item Quantized networks with TVM ~\cite{DBLP:journals/corr/abs-2006-10226} 
\end{itemize}

\section{Symmetric Quantization}
\label{sec:symmetric_quant}

The most common quantization approach represents floating point 32-bit ($fp32$) numbers with integer 8-bit ($int8$ or $uint8$) numbers ~\cite{pete_warden, DBLP:journals/corr/abs-1712-05877, DBLP:journals/corr/WuLWHC15}, which reduces the memory footprint by a factor of four. Furthermore, 8-bit integer operations can be executed highly efficiently (low power and high performance) on machine learning accelerator hardware. 

For symmetric quantization, where the zero offset is zero, equation \ref{eqn0} expresses the relationship between quantized tensor ${X_q}$ and original $fp32$ tensor ${X}$. 

\begin{equation} \label{eqn0}
X = scale_X * X_q
\end{equation}

There are multiple ways to determine the $scale_X$ in equation \ref{eqn0}. One approach might be to profile the $fp32$ tensor to determine the maximum numerical range and mapping this range to the full $int8$ range. Another might be to minimize the overall quantization error by creating profile histograms and saturating the numerical range prior to mapping. Precisely, this is one of the motivations for this paper, i.e. decoupled quantized model development from the target hardware platform and its compiler. Giving the scale, and the data type of the quantized tensor (i.e. $int8$ or $uint8$), the quantized tensor $X_q$ can be calculated as ${1\over scale_X} * X$, with additional rounding and clipping stage to insure that the quantized tensor is represented as proper $int8$ or $uint8$ values.

Similarly, to quantizing individual tensors, each layer of a network needs to be quantized. The fully connected layer, i.e. a matrix-matrix-multiply followed by addition of a bias tensor, serves as an example here.
The fully connected layer with input tensor $X$, weight $W$, bias $B$ and output tensor $Y$ is given in equation \ref{eqn1}.
\begin{equation} \label{eqn1}
Y = W \cdot X + B
\end{equation}
\begin{equation} \label{eqn2}
Y_q = {{(scale_W * scale_X)}\over {scale_Y} } * ( W_q \cdot X_q + B_q)
\end{equation}
\begin{equation} \label{eqn3}
Y_q= {{(scale_W * scale_X)}\over {scale_Y} } * Y_{intermediate}
\end{equation}
The intermediate tensor $ Y_{intermediate}$ is of $INT32$ data type and is the result of the MatMulInteger of the quantized weight and quantized input tensor with the integer addition of the quantized bias as given in equation \ref{eqn4}.
\begin{equation} \label{eqn4}
Y_{intermediate} = W_q \cdot X_q + B_q
\end{equation}
The bias $B$ is quantized to be of same scale as the output of the MatMulInteger operation and is given as $INT32$ value.
\begin{equation} \label{eqn5}
B_q= {1\over {scale_W * scale_X}} * B
\end{equation}
The ${{(scale_W * scale_X)}\over {scale_Y} }$ represents the rescaling (aka output quantization) of the fully connected layer and is used in equation \ref{eqn3} to determine the quantized layer output. Just as for the individual tensor above, a rounding and clipping stage follows equation \ref{eqn3} to ensure that the output tensor is represented as proper $int8$ or $uint8$ values.
Similarly, convolution layers will include such rescaling stage.
\subsection{Rescaling}
\label{sec:rescale}

As shown in the previous section for fully connected layer, a rescaling with rounding and clipping is required after certain network layers. The rescaling values are floating point values, which can be greater or smaller than 1. To perform the rescaling with integer arithmetic, the floating-point multiplication is replaced with an integer multiplication by an integer value followed by a bitwise right shift. A right shift by $N$ bits is equivalent to a division by $2^N$. Utilizing 2 Mul operators both the integer value and the number of right shift bits can be codified within the ONNX network. 

\par
\noindent
\textbf{Method with 2 Mul operators}
\begin{enumerate}
\itemsep0em
\item{${Quant\_scale}$ is an Integer value represented as FLOAT.}
\item{$Quant\_shift = {1\over 2^N}$. representing a right shift by $N$ bits}
\end{enumerate}

Alternatively, only the floating-point scaling value is codified in ONNX and the conversion to integer value and number right shifts is the responsibility of the hardware-specific tool chain.

\par
\noindent
\textbf{Method with 1 Mul operator}
\begin{enumerate}
\itemsep0em
\item{$Quant\_multiplier = Quant\_scale * Quant\_shift $ }
\end{enumerate}

For example, a $Quant\_multiplier$ of 0.25 can be represented by $Quant\_scale$ of 1 and $Quant\_shift$ of $1\over{2^2}$. 
A $Quant\_multiplier$ of ${1\over{3}}$ can be represented by $Quant\_scale$ of 11184810 and $Quant\_shift$ of $1\over{2^{25}}$.
It should be noted, as the $Quant\_scale$ integer value is stored as FLOAT, the largest exactly represented integer value is ${2^{24}} = 16,777,216$

The rounding and clipping that follows rescaling is codified in the pre-quantized ONNX network with the ONNX operator QuantizeLinear with $(scale=1,zero\_point=0)$. per QuantizeLinear API, the data type of the $zero\_point$ argument determines the data type of the output tensor. I.e., an $int8$ $zero\_point$ argument results in $int8$ output, while an $uint8$ $zero\_point$ arguments results in $uint8$ output. Here, the QuantizedLinear is not used for rescaling, and $scale$ is set to 1, as the scaling has already been codified using one or two MUL operators.

\section{Fully Connected Layer}
\label{sec:FC}

\begin{figure}[htb]
\includegraphics[width=\columnwidth]{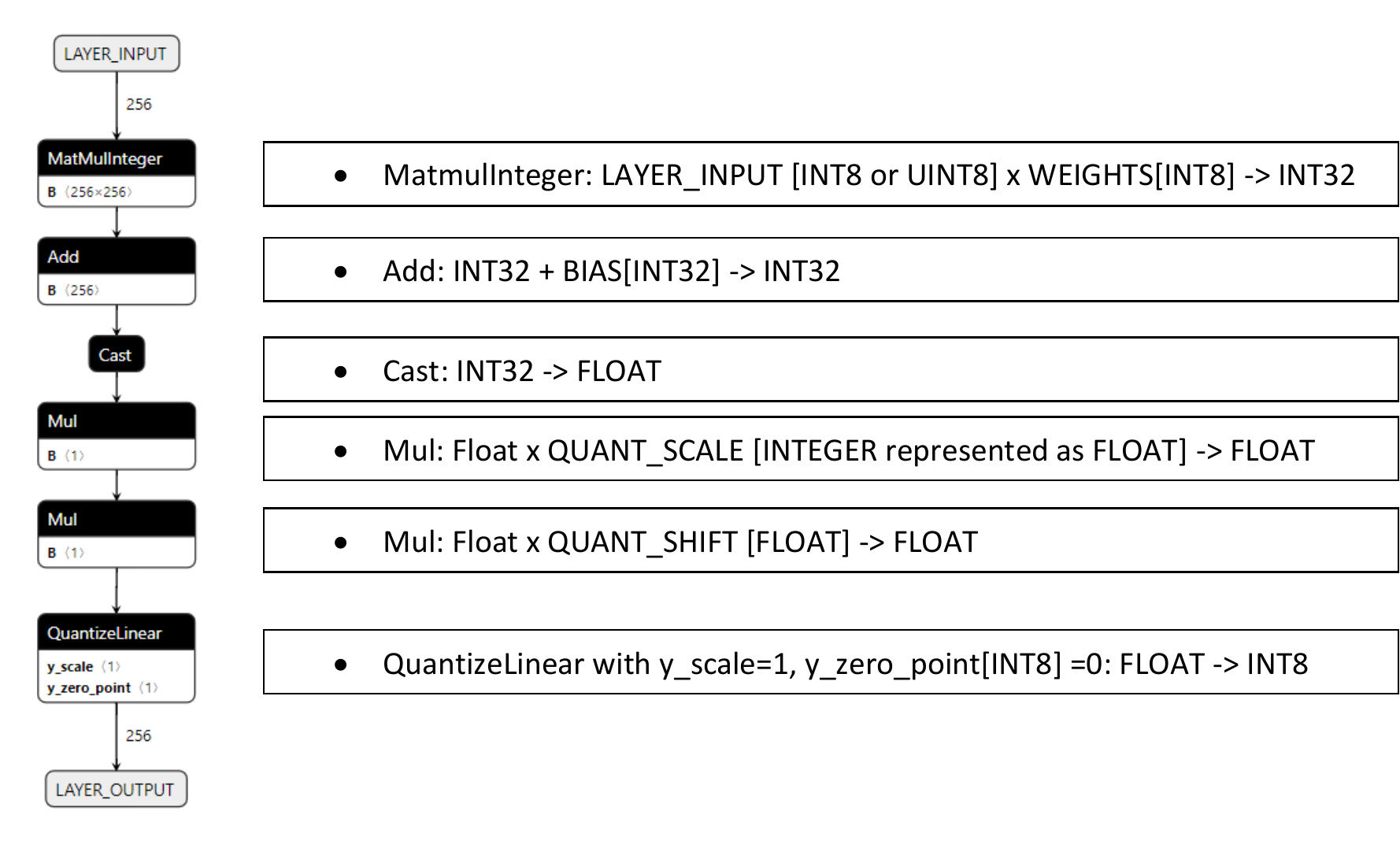}
\caption{Fully Connected Layer without activation function. Rescaling expressed with two Mul operations.}
\label{fig:fc_no_activation}
\end{figure}

\begin{figure}[htb]
\includegraphics[width=\columnwidth]{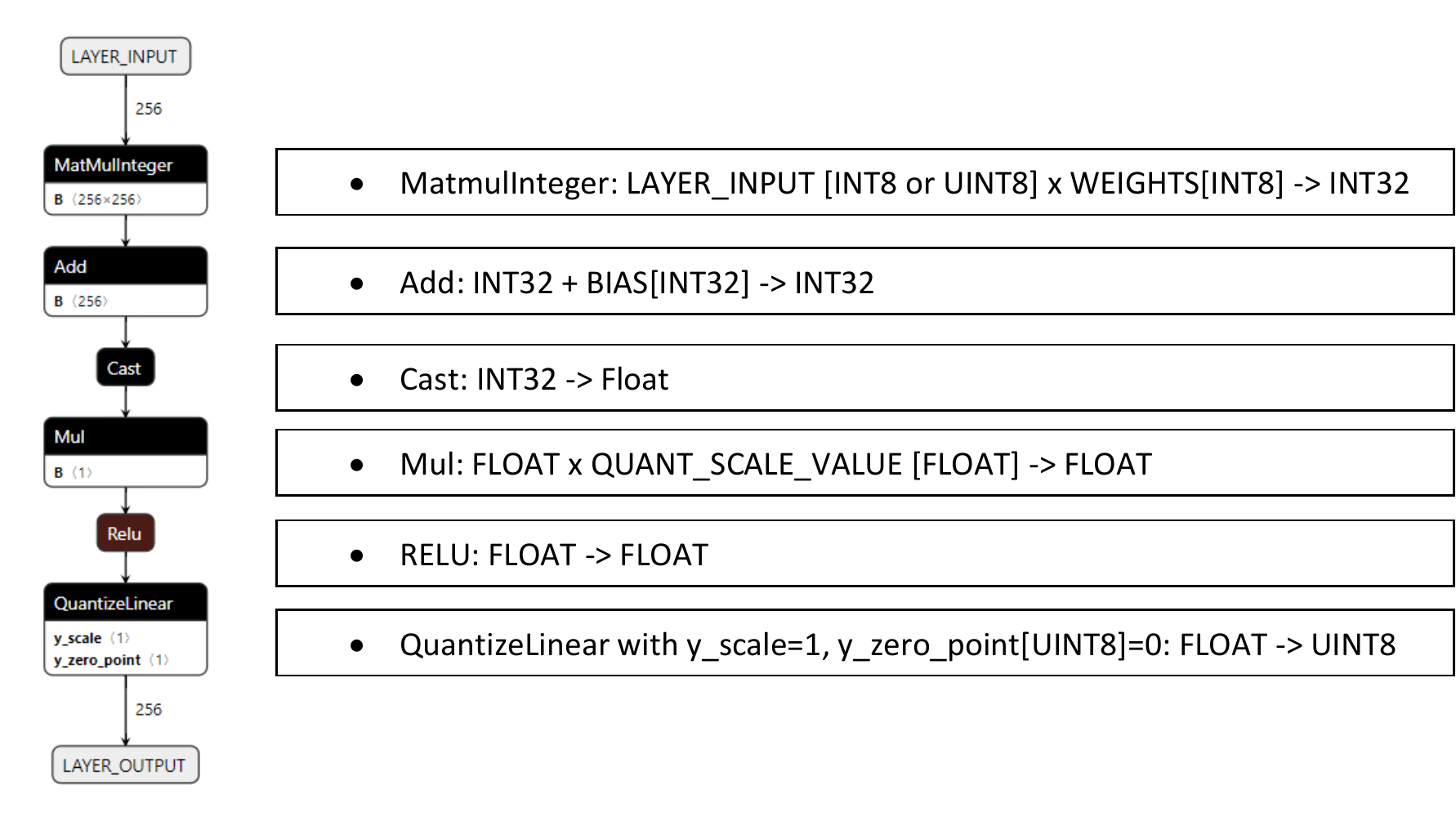}
\caption{Fully Connected Layer with ReLU activation function. Rescaling expressed with one Mul operation.}
\label{fig:fc_with_relu}
\end{figure}

As a first example we show the fully connected layer of a Multi Layer Perceptron (MLP) network. The example shows the methodology applied to a complete network with input and output that can be run within the ONNXruntime. This is the base pattern that can be applied to larger MLP models and other networks containing fully connected layers. Fig. \ref{fig:fc_no_activation} shows the ONNX flow for a fully connected layer without an activation function, while Fig. \ref{fig:fc_with_relu} shows the layer with ReLU activation. In these figures, the ONNX graphs are visualized using Netron ~\cite{netron} tool on the left and the individual operator steps are shown on the right. For each operator, the data types of it’s input and output tensors are given. 
The ONNX operator MatMulInteger is used to express the matrix-matrix-multiply of the layer input of type $int8$ or $uint8$, with the weight coefficients given as $int8$ resulting in an output tensor of type $int32$. Following the MatMulInteger, the bias, as $int32$ data type is added using the ONNX Add operator. The rescaling is expressed here using two ONNX Mul operator with $fp32$ inputs, thus a ONNX Cast operator is added to cast the $int32$ into $fp32$. The final stage in this pattern is the rounding and clipping performed by the ONNX QuantizeLinear operator.

\section{Convolution Layer}
\label{sec:Conv}

\begin{figure}[htb]
\includegraphics[width=\columnwidth]{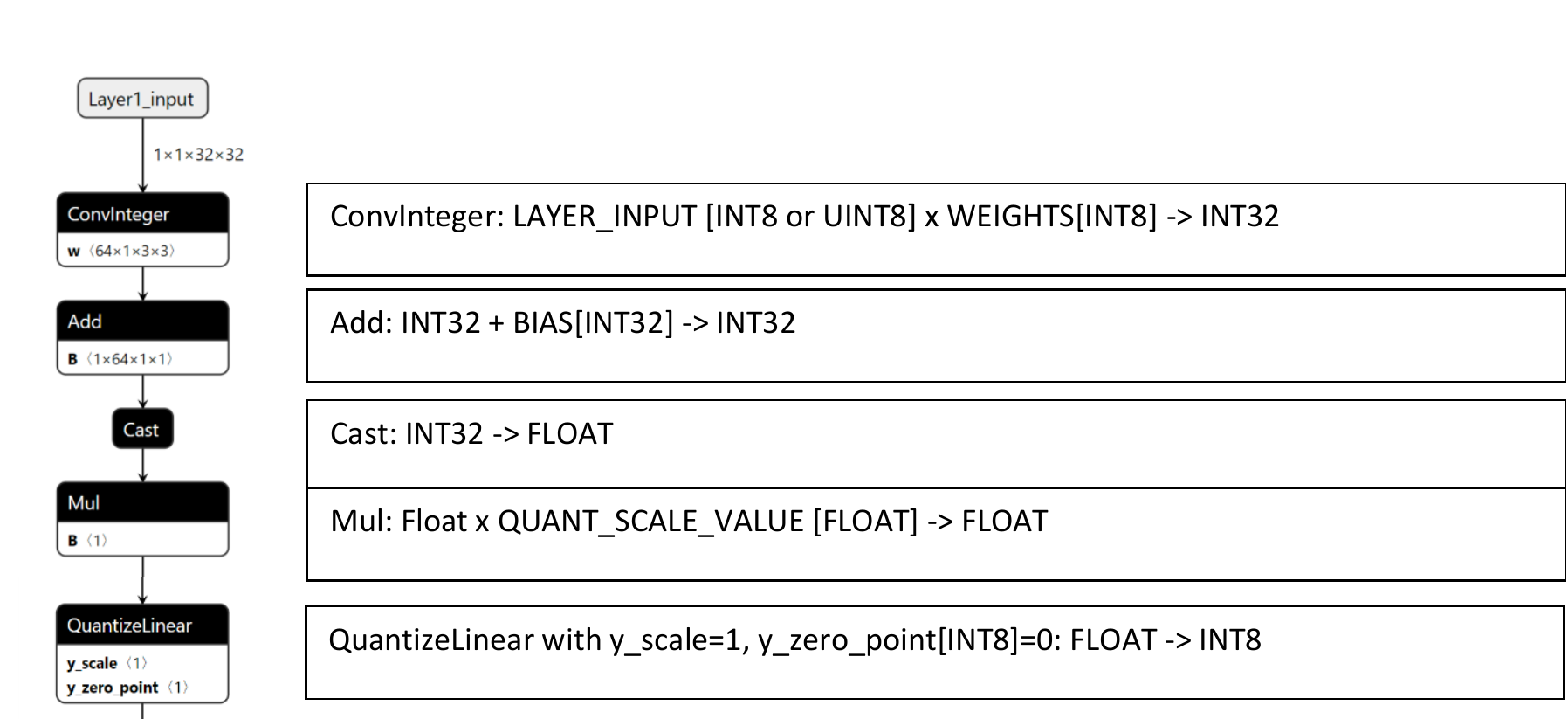}
\caption{Convolution Layer without activation function. Rescaling expressed with one Mul operation.}
\label{fig:conv}
\end{figure}

As second example we show the Convolution 2D layer of a Convolutional Neural Network (CNN). The methodology applied to a complete network with input and output that can be run within the ONNXruntime. Fig. \ref{fig:conv} shows a Convolution layer without activation function. A ReLU activation function will be similarly handled as for the Fully Connected layer shown in Fig. \ref{fig:fc_with_relu}. The ONNX operator ConvInteger is used to express the Convolution with the kernel coefficients given as $int8$ weights. Following the convolution, the bias, as $int32$ data type is added using the ONNX Add operator. The rescaling is expressed using the ONNX Mul operator with $fp32$ inputs, thus a ONNX Cast operator is added to cast the $int32$ into $fp32$. The final stage in this pattern is the rounding and clipping performed by the ONNX QuantizeLinear operator.

\section{Tanh and Sigmoid activation functions within an $int8$ quantized network}
\label{sec:fp16 activations}
\begin{figure}[h!]
%\begin{figure}[htb]
\includegraphics[width=\columnwidth]{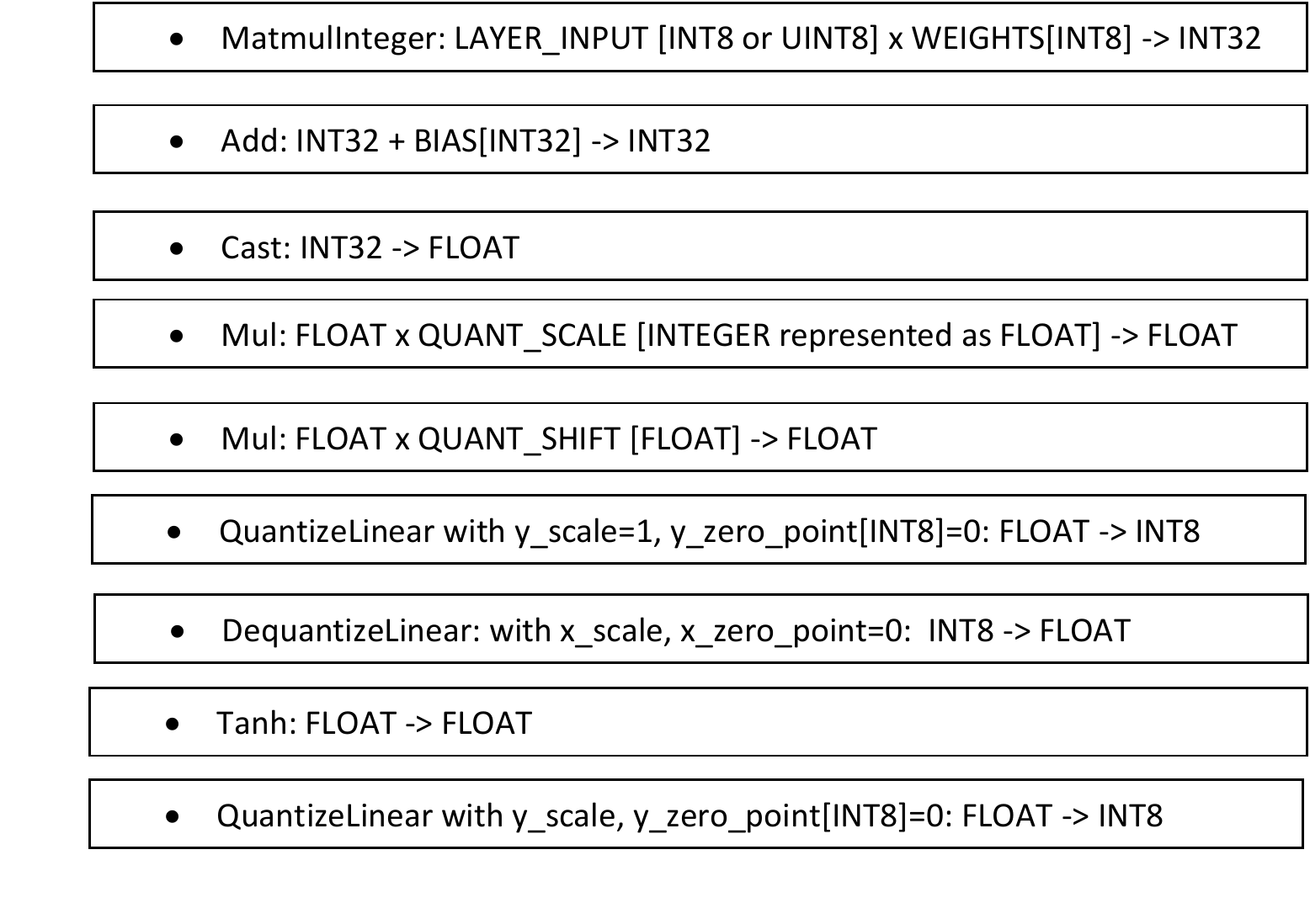}
\caption{Fully Connected Layer with $int8$ Tanh activation function. Rescaling expressed with two Mul operations. }
\label{fig:fc_with_int8_tanh}
\end{figure}

The above described methodology is further expanded to Tanh and Sigmoid activation functions. The figures in this section show the individual operator steps in detail only and omit the ONNX graph visualization.
Fig. \ref{fig:fc_with_int8_tanh} shows the sequence of ONNX operators to codify an $int8$ quantized network with $int8$ tanh activation function. The $Quant\_scale$ and $Quant\_shift$ values are set such that the full input range of tanh is mapped to the quantized $int8$ range. The $y\_scale$ is determined by mapping the $int8$ range to the full output range of tanh. It should be noted that the ONNX operator for tanh is for $fp32$ input resulting in $fp32$ output. Setting the rescale to map to full input range and the $y\_scale$ in the above described way results in using a $int8$ tanh approximation.

\begin{figure}[h!]
%\begin{figure}[htb]
\includegraphics[width=\columnwidth]{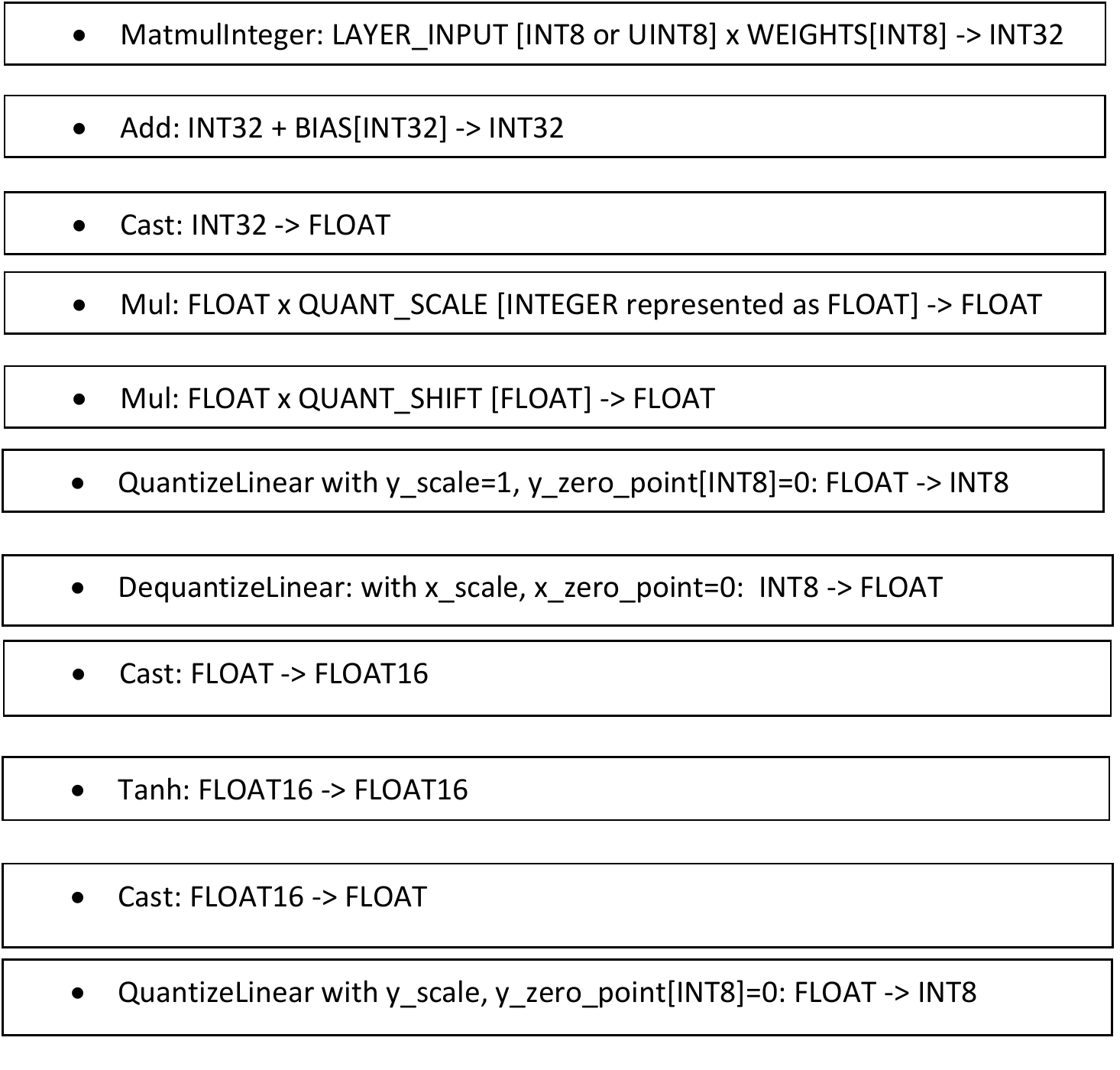}
\caption{Fully Connected Layer with $fp16$ Tanh activation function. Rescaling expressed with two Mul operations. }
\label{fig:fc_with_fp16_tanh}
\end{figure}

\begin{figure}[h!]
%\begin{figure}[htb]
\includegraphics[width=\columnwidth]{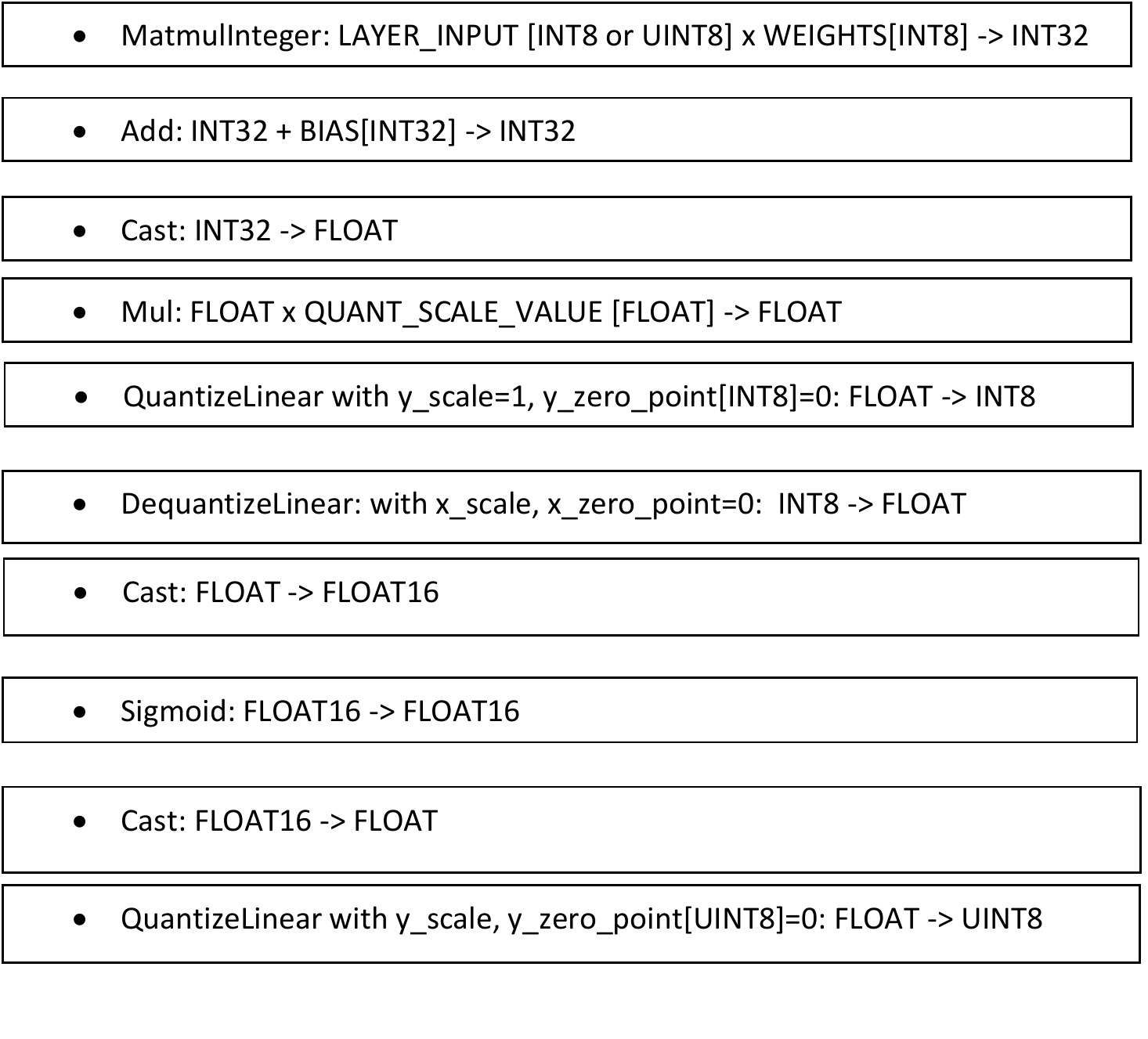}
\caption{Fully Connected Layer with $fp16$ Sigmoid activation function. Rescaling expressed with one Mul operation. }
\label{fig:fc_with_fp16_sigmoid}
\end{figure}

An alternative approach is to perform tanh or sigmoid as floating-point functions. These functions might be implemented as $fp32$ or, as shown in the following examples, as $fp16$.
Fig. \ref{fig:fc_with_fp16_tanh} shows a mixed $int8$/$fp16$ flow which allows rescaling to a narrow input range (symmetric around zero) of tanh and execution of tanh function in $fp16$ precision.
Fig. \ref{fig:fc_with_fp16_sigmoid} shows the mixed $int8$/$fp16$ flow for the sigmoid activation function. As the sigmoid activation produces always positive outputs, the quantized output will be $uint8$.

\section{Conclusion}
\label{sec:conclusion}

This paper presents a methodology to separate the quantization process from the model compilation stage which enables independent development, while allowing hardware/software co-design. Detailed examples are given for both MLP and CNN based networks, which can be extended to other networks in a straightforward fashion.

%\clearpage
 
{\footnotesize
\bibliography{joint_pq_paper}}
\bibliographystyle{IEEEtran}

\end{document}